\documentclass[10pt,twocolumn,letterpaper]{article}

\usepackage[pagenumbers]{cvpr} % To force page numbers, e.g. for an arXiv version

\usepackage{url}            %
\usepackage{booktabs}       %
\usepackage{amsfonts}       %
\usepackage{nicefrac}       %
\usepackage{microtype}      %
\usepackage{xcolor}         %
\usepackage{multirow}
\usepackage{graphicx}
\usepackage{enumitem} %
\usepackage{amsmath}

\usepackage{caption} %
\usepackage{float} %
\usepackage{bm}

\usepackage{subcaption}
\usepackage{tabularx}
\usepackage{makecell}
\usepackage{pifont}

\usepackage{enumitem}
\usepackage{bm}
\usepackage{url}
\usepackage{comment}
\usepackage{graphicx}
\usepackage{colortbl}
\usepackage{xcolor}
\usepackage{amsmath}
\usepackage{multirow}
\usepackage{adjustbox}
\usepackage{threeparttable}
\usepackage{epsfig}
\usepackage{wrapfig}
\usepackage{algorithm}
\usepackage{algorithmic}
\usepackage{setspace}
\usepackage{bm}
\usepackage{float}
\usepackage{makecell}

\usepackage{booktabs}       % professional-quality tables
\usepackage{amsfonts}       % blackboard math symbols
\usepackage{nicefrac}       % compact symbols for 1/2, etc.
\usepackage{microtype}      % microtypography
\usepackage{tablefootnote}

\definecolor{cvprblue}{rgb}{0.21,0.49,0.74}
\usepackage[pagebackref,breaklinks,colorlinks,allcolors=cvprblue]{hyperref}
\usepackage[table,xcdraw]{xcolor}
\definecolor{row_color}{HTML}{E0FFFF} % 浅青色

\def\paperID{} % *** Enter the Paper ID here
\def\confName{CVPR}
\def\confYear{2026}

\title{IDESplat: Iterative Depth Probability Estimation for \\ Generalizable 3D Gaussian Splatting}
\author{
Wei Long\textsuperscript{1}, Haifeng Wu\textsuperscript{1}, Shiyin Jiang\textsuperscript{1}, Jinhua Zhang\textsuperscript{1}, Xinchun Ji\textsuperscript{2}, Shuhang Gu\textsuperscript{1*} \\
\textsuperscript{1}University of Electronic Science and Technology of China \\
\textsuperscript{2}Aerospace Information Research Institute, Chinese Academy of Sciences \\
{\tt\small lwsch5940@163.com, shuhanggu@gmail.com} \\
}

\begin{document}

\twocolumn[{
\maketitle

\vspace{-7mm}\hspace{-2.5mm}
\centerline{
\includegraphics[width=0.98\linewidth,trim={4pt 4pt 4pt 4pt}]{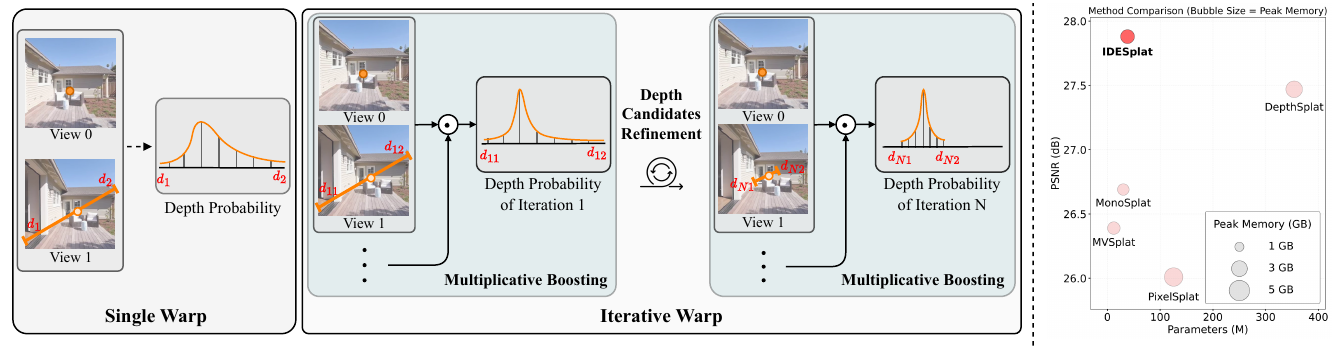}}
\vspace{-1mm}
\captionof{figure}
{
\textbf{Left:} Methods~\cite{chen2024mvsplat, xu2025depthsplat, liu2025monosplat} that estimate depth probability via a single warp operation. \textbf{Middle:} Our IDESplat can iteratively leverage multi-warp operations to boost the depth probability estimate and refine the depth candidates for accurate Gaussian mean predictions.
\textbf{Right: }The experimental results of our IDESplat compared with mainstream methods such as PixelSplat~\cite{charatan2024pixelsplat}, MVSplat~\cite{chen2024mvsplat}, MonoSplat~\cite{liu2025monosplat}, and DepthSplat~\cite{xu2025depthsplat}. The PSNR values are reported for the entire RE10K test set.
% \small
% \textbf{Vanilla next-scale prediction \textit{vs}. MVAR.}
% \textbf{\small Left}: The vanilla next-scale method predicts each scale based on all previous scales and requires each token to consider all preceding tokens. 
% \textbf{\small Middle}: A next-scale variant predicts the next scale using only the adjacent scale, leveraging scale Markovian conditioning across scales.
% \textbf{\small Right}: MVAR further disentangles the spatial constraint and predicts each token using the neighboring positions of adjacent scales, based on spatial Markovian conditioning.
}
\label{fig:intro}
% \vspace{2em}
\vspace{3mm}
% \maketitle
}]
% \maketitle

% 添加通讯作者信息
\renewcommand{\thefootnote}{\fnsymbol{footnote}}
\footnotetext[1]{corresponding author}

\begin{abstract}
Generalizable 3D Gaussian Splatting aims to directly predict Gaussian parameters using a feed-forward network for scene reconstruction.
Among these parameters, Gaussian means are particularly difficult to predict, so depth is usually estimated first and then unprojected to obtain the Gaussian sphere centers.
Existing methods typically rely solely on a single warp to estimate depth probability, which hinders their ability to fully leverage cross-view geometric cues, resulting in unreliable and coarse depth maps.
To address this limitation, we propose \textbf{IDESplat}, which iteratively applies warp operations to boost depth probability estimation for accurate Gaussian mean prediction.
First, to eliminate the inherent unreliability of a single warp, we introduce a Depth Probability Boosting Unit (DPBU) that integrates epipolar attention maps produced by cascading warp operations in a multiplicative manner.
Next, we construct an iterative depth estimation process by stacking multiple DPBUs, progressively identifying potential depth candidates with high likelihood.
As IDESplat iteratively boosts depth probability estimates and updates the depth candidates, the depth map is gradually refined, resulting in accurate Gaussian means.
%
% Finally, for the other Gaussian parameters, we design a Gaussian Focused Module (GFM) to determine the most relevant Gaussian tokens for feature interaction.
%
We conduct experiments on RealEstate10K, ACID and DL3DV.
IDESplat achieves outstanding reconstruction quality and state-of-the-art performance with real-time efficiency.
On RE10K, it outperforms DepthSplat by \textbf{0.33 dB} in PSNR, using only \textbf{10.7\%} of the parameters and \textbf{70\%} of the memory.
Additionally, our IDESplat improves PSNR by \textbf{2.95 dB} over DepthSplat on the DTU dataset in cross-dataset experiments, demonstrating its strong generalization ability.
The code is available at {\href{https://github.com/CVL-UESTC/IDESplat}{https://github.com/CVL-UESTC/IDESplat}}.

\end{abstract}

\section{Introduction}
\label{sec:intro}

Single-scene 3D Gaussian Splatting (3DGS) benefits from a rasterization-friendly pipeline, making it well-suited for real-time scene reconstruction~\cite{kerbl20233d, yu2024mip}, but it suffers from limited generalization capabilities.
Generalizable 3DGS~\cite{charatan2024pixelsplat, chen2024mvsplat, liu2024mvsgaussian, xu2025depthsplat, zhang2024gaussian} addresses this shortcoming by using a feed-forward network to directly predict all Gaussian parameters, enabling it to handle unseen scenes.
Among all the Gaussian sphere parameters, the Gaussian mean is crucial but difficult to predict directly due to the local support nature of Gaussian gradients~\cite{charatan2024pixelsplat}, which significantly affects the optimization of the overall parameters.
Existing methods~\cite{chen2024mvsplat, liu2025monosplat, zhang2025transplat, xu2025depthsplat} commonly require first estimating the pixel-wise depth map and then unprojecting it to obtain the centers of the Gaussian spheres.
% 
% This design reduces the difficulty of the network optimization process while increasing the reliance on accurate depth estimation.
%
This design reduces the difficulty of the network optimization process by decoupling the prediction of Gaussian mean parameters, while increasing the reliance on accurate depth estimation.

Efforts have been made to obtain accurate and refined depth estimation.
Early methods~\cite{charatan2024pixelsplat, szymanowicz2024splatter} directly use differentiable operations to predict the depth probability distribution from image features.
Although these methods can predict the depth map for scene reconstruction in a generalizable way, their ability to exploit multi-view feature similarity is limited, which restricts their performance.
Subsequent approaches~\cite{chen2024mvsplat, liu2024mvsgaussian} introduce cost volumes that use warp operations to establish feature similarity across views, which provide valuable geometric cues for depth estimation and simplify the network learning process.
However, these methods rely solely on a single warp to model feature similarity, which prevents them from fully exploiting the rich geometric information, leading to unreliable and coarse depth maps.
% 
% Therefore, how to incorporate multiple warps to mine rich geometric information for cross-view similarity calculations, in order to produce refined and reliable depth maps for accurate Gaussian means, remains a key challenge.
%
% Therefore, how to incorporate multiple warps to gradually leverage rich cross-view details, thereby producing refined and reliable depth maps, remains a key challenge.
%
%
Therefore, how to incorporate multiple warps to gradually leverage rich cross-view geometric details, producing refined and reliable depth maps for accurate Gaussian mean prediction, remains a key challenge.
%
% Therefore, how to incorporate multiple warps to gradually leverage rich cross-view geometric details in a memory-efficient manner, thereby producing refined and reliable depth maps for accurate Gaussian mean prediction, remains a key challenge.

%
In this paper, we propose IDESplat, which iteratively performs warps to refine depth maps for accurate Gaussian means prediction.
By integrating cascade warp results, we can progressively boost the feature similarity measure to identify high-likelihood surface points and suppress low-probability depth candidates.
The iterative warp framework of our IDESplat is shown in Fig.~\ref{fig:intro}.
%
%Firstly, to eliminate the unreliability of a single warp, we introduce a Depth Probability Boosting Unit (DPBU) to fuse multi-level epipolar attention maps for reliable depth map estimation.
%
Firstly, to eliminate the unreliability of a single warp, we introduce a Depth Probability Boosting Unit (DPBU) to fuse multiple epipolar attention maps in a multiplicative manner for a reliable depth map.
Next, we gradually update the depth search range while increasing the feature resolution, enabling warp and correlation calculations at a finer scale.
Feature matching becomes easier and more precise as the depth search range is re-centered and image features are enhanced in this process.
Finally, for other Gaussian parameters, we propose a Gaussian Focused Module that determines the most relevant Gaussian tokens to compute attention weights for feature interaction.
The experimental results on RealEstate10K show that IDESplat outperforms DepthSplat by \textbf{0.33 dB} in PSNR with only \textbf{10.7\%} of the parameters and \textbf{70\%} of the memory. 
%
% Compared to MonoSplat, IDESplat achieves a \textbf{1.12 dB} improvement in PSNR with a slight increase in parameters.
%
Moreover, IDESplat shows outstanding cross-dataset generalization, achieving 28.79 dB on ACID when transferred from RE10K, outperforming methods trained directly on ACID. It also improves PSNR by \textbf{2.95 dB} over DepthSplat on the DTU dataset in cross-dataset experiments.
% %
In summary, the main contributions of this paper are as follows:
%
% 待修改
\begin{itemize}
\item We propose IDESplat, a generalizable feedforward 3DGS model, which iteratively performs warps to progressively boost the feature similarity measure and refine depth maps for accurate Gaussian mean prediction.
\item To eliminate the inherent unreliability of a single warp, we introduce a Depth Probability Boosting Unit that multiplicatively integrates multiple epipolar attention maps for reliable and refined depth map estimation.
\item We design a Gaussian Focused Module to identify the most relevant Gaussian tokens for computing attention scores and reweight enhanced features.
\item Experiments on RealEstate10K ACID, and DL3DV show that our IDESplat significantly improves reconstruction quality and generalization capability while maintaining real-time inference efficiency.
\end{itemize}

\section{Related Work}
\label{sec:related_work}

\noindent \textbf{Generalizable 3D Gaussian Splatting.}
Single-scene 3D Gaussian Splatting methods~\cite{kerbl20233d, fan2024instantsplat, yan2024street, niemeyer2025radsplat, feng2025flashgs} enable more efficient rendering than neural fields and volume rendering methods~\cite{lombardi2019neural, sitzmann2020implicit, mildenhall2021nerf, xie2022neural} thanks to their rasterization-friendly formulation, but they still suffer from long optimization time and limited generalization.
To address this issue, generalizable 3D Gaussian Splatting methods~\cite{charatan2024pixelsplat, tang2024hisplat, liu2025monosplat, xu2025depthsplat} predict all Gaussian parameters in a single feed-forward pass, enabling fast reconstruction and novel view synthesis for unseen scenes.
Early works~\cite{charatan2024pixelsplat, szymanowicz2024splatter} directly regressed Gaussian parameters from image features using feed-forward networks.
With the introduction of cost-volume construction into generalizable 3DGS, subsequent methods~\cite{chen2024mvsplat, liu2024mvsgaussian} exploited cross-view geometric cues to improve depth estimation and simplify Gaussian parameter prediction.
More recently, methods~\cite{xu2025depthsplat, liu2025monosplat} further improved performance by incorporating pre-trained monocular depth models.
However, existing methods still rely on a single warp for depth estimation, which limits their ability to fully exploit cross-view feature cues.
In contrast, IDESplat integrates feature similarity from cascaded warps to produce more reliable and refined depth maps for Gaussian mean prediction.

\noindent \textbf{Iteration-based Optimization Methods.}
\noindent Iteration-based optimization methods~\cite{adler2017solving, li2018deepim} are widely used in various tasks~\cite{he2024lotus, zhu2024addressing, chen2024virtual} due to their ability to progressively enhance and refine the feature learning process.
For instance, in optical flow estimation~\cite{teed2020raft, hui2018liteflownet,  yang2019volumetric, sun2018pwc, ilg2017flownet}, iterative optimization is commonly used to build coarse-to-fine pyramidal features by stacking multiple feature units, resulting in more accurate and stable flow estimation.
In the field of depth estimation, some works~\cite{lipson2021raft, wang2022itermvs, wang2022efficient, ma2022multiview, xu2023iterative, xu2025igev++, chen2024mocha} have iteratively retrieved multi-view correlation features using structures like GRUs and LSTMs with shared parameters to update the disparity field, yielding better depth estimation results.
Similarly, in monocular depth estimation~\cite{zuo2024ogni, shao2023iebins, fu2018deep}, iterative methods are often used to gradually refine the depth search range, leading to more stable and accurate predictions.
Additionally, in the 3D Gaussian Splatting (3DGS) domain, recent methods~\cite{wenlife, zhang2025transplat, xu2025resplat, chen2024g3r, harrison2022closer} have also attempted to design iterative optimization processes to reduce the difficulty of 3D Gaussian reconstruction tasks.
Unlike existing iterative optimization methods, we propose a novel Depth Probability Estimation Unit that integrates the similarity results of multiple warps in a multiplicative manner.
This approach produces more reliable and refined depth maps, while progressively enhancing feature resolution and refining the depth candidate range throughout the iterative process.

\section{Method}
\label{sec:method}
\begin{figure*}
    \centering
    \includegraphics[width=\linewidth]{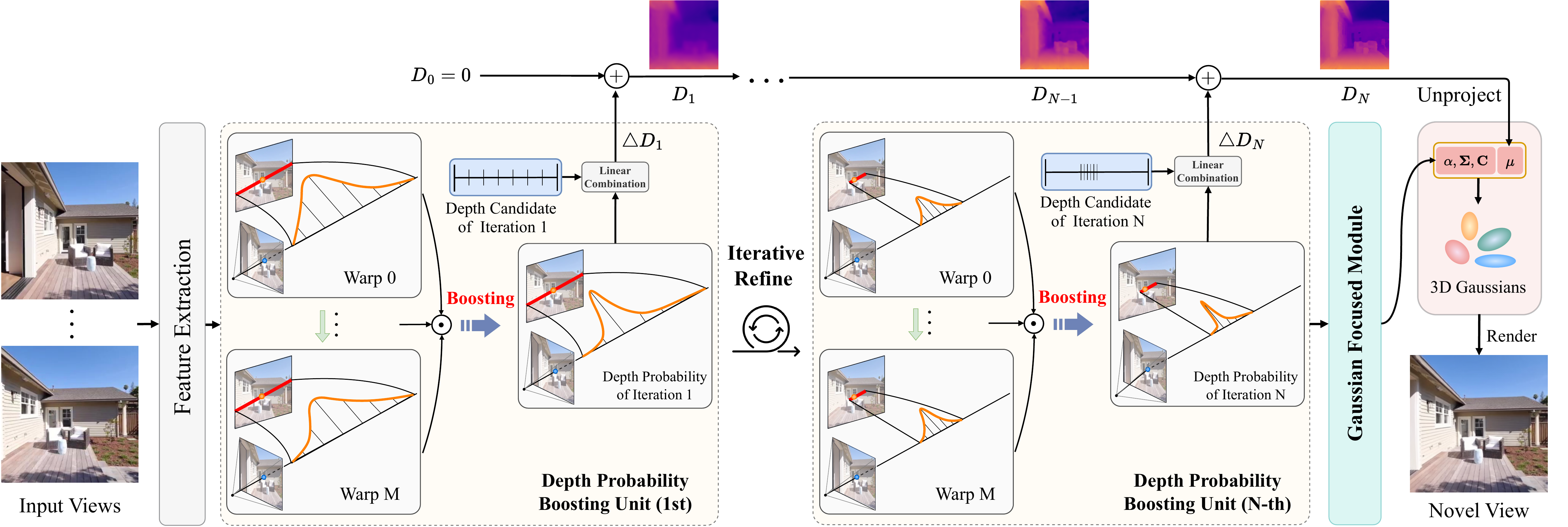}
    \caption{The overall architecture of IDESplat. IDESplat consists of three key parts: a feature extraction backbone, an iterative depth probability estimation process, and a Gaussian Focused Module (GFM). The iterative process is built upon cascaded Depth Probability Boosting Units (DPBUs). Within each DPBU, depth probabilities are estimated sequentially by stacked Warp-Index Epipolar Attention (WIEA) blocks, where each block operates on progressively refined features and thus produces distinct probability estimates. Each unit combines multi-level warp results in a multiplicative manner to mitigate the inherent unreliability of a single warp. As IDESplat iteratively updates the depth candidates and boosts the probability estimates, the depth map becomes more precise, leading to accurate Gaussian means.}
    \label{fig:IDESplat}
    \vspace{-4mm}
\end{figure*}

\subsection{Preliminaries.}
Given a sequence of $V$ sparse-view images $\mathcal{I} = \{ {\mathcal I}^{i}\}_{i=1}^{V}$, where each image ${\mathcal I}^i$ has dimensions $H \times W \times 3$, and the corresponding camera projection matrices are ${\mathcal P}^i \in \mathbb{R}^{3 \times 4}$.
The goal of the generalizable 3DGS task is to learn a network that maps images to 3D Gaussian parameters. This process is defined as:
\begin{equation}
\label{eq:eq1}
    f_{\bm \theta}: \{ ({\mathcal I}^{i}, {\mathcal P}^i) \}_{i=1}^{V} \mapsto \{ (\bm{\mu}_j, \bm{\alpha}_j, \bm{\Sigma}_j, \bm{c}_j ) \}_{j=1}^{V \times H \times W},
\end{equation}
where $f_{\bm \theta}$ is a feed-forward network with learnable parameters ${\bm \theta}$.
The parameters $\bm{\mu}_j$, $\bm{\alpha}_j$, $\bm{\Sigma}_j$, and $\bm{c}_j$ are the Gaussian mean, opacity, covariance, and color, respectively.
%
% With one Gaussian corresponding to each pixel, the total number of Gaussians equals the total number of pixels across all input images.
%
Since Gaussian means are difficult to predict directly, most existing methods estimate them by unprojecting a depth map into 3D.
The depth estimation scheme is as follows:
First, the input image sequence is processed using a multi-view feature extraction backbone, resulting in a downsampled feature $\bm F \in \mathbb{R}^{V \times \frac{H}{4} \times \frac{W}{4} \times C}$.
%
% Next, a discrete set of depth candidates is defined as ${\bm G} = [d_1, d_2, \cdots, d_D] \in \mathbb{R}^{D}$.
%
Then, the cost volume is computed based on $\bm{F}$, the camera projection matrices ${\mathcal{P}}^i$, ${\mathcal{P}}^j$ for different views, and the depth candidates ${\bm{G}} = [d_1, d_2, \cdots, d_D] \in \mathbb{R}^{D}$.
Specifically, the feature ${\bm F}^j$ from view $j$ is warped to view $i$ as follows:
\begin{equation}
\label{eq:warp_ori}
    {\bm F}^{j \to i} = \mathcal{W}({\bm F}^j, {\mathcal P}^i, {\mathcal P}^j, \bm G) \in \mathbb{R}^{\frac{H}{4} \times \frac{W}{4}  \times D\times C},
\end{equation}
where $\mathcal{W}$ denotes the warp operator. The correlation is then computed as the dot product between $\bm{F}^{i}$ and $\bm{F}^{j \to i}_{d_m}$:
\begin{equation}
\label{eq:correlation}
    {\bm C}_{d_m}^{i} = ({\bm F}^i \cdot {\bm F}^{j \to i}_{d_m} / \sqrt{C}) \in \mathbb{R}^{\frac{H}{4} \times \frac{W}{4}},
\end{equation}
where $m \in \{1,\ldots,D\}$ and $C$ denotes the feature channel dimension. 
%
% This typical correlation computation method has to store all the dense warping features, which leads to a memory issue, especially when the depth candidate size $D$ is large.
%
Subsequently, $\bm{C}^{i} = [\bm{C}_{d_1}^{i}, \ldots, \bm{C}_{d_D}^{i}]$ is refined with a U-Net and upsampled to the input resolution as $\tilde{\bm{C}}^{i} \in \mathbb{R}^{H \times W \times D}$.
Finally, a $\mathrm{softmax}$ function is applied to obtain the probability of each candidate depth, and the final depth map is computed as their weighted average.

To achieve high-quality 3D reconstruction, an accurate depth map $\bm{D}$  is crucial since it determines the centers of 3D Gaussians.
%
% So, the central challenge is how to fully leverage multi-view features to accurately estimate depth probabilities at the candidate depths.
%
However, this commonly used method only relies on a single cross-view warp to compute feature similarity, underutilizing cross-view geometry and often leading to unreliable, coarse depth estimates.
In addition, this typical correlation computation method requires storing all the dense warping features, which incurs significant memory overhead, especially when the depth candidate size $D$ is large.

\subsection{Iterative Multi-View Depth Estimation}
\label{sec:i_d_e}

In this paper, we propose IDESplat to iteratively boost the feature similarity measure using multiple warp operations.
As IDESplat progressively mines geometric cues to identify high-probability potential depth candidates, it can produce refined and reliable depth probability estimation results for precise Gaussian mean prediction.

\noindent \textbf{Warp-Index Epipolar Attention.}
The warp operation is key to constructing the cost volume or epipolar attention map, as it establishes pixelwise correspondences between features in the source and target views.
However, as shown in Eq.~\eqref{eq:warp_ori}, this warping operation requires sampling target-view features for each depth candidate, thereby incurring high memory cost.
To alleviate the inherent memory overhead, we introduce a Warp-Index Epipolar Attention mechanism that only stores warp indices for similarity matrix multiplication.
We first compute the warping index map $\bm{I}$ as follows:
\begin{equation}
\label{eq:warp}
    {\bm I}^{j \to i} = \mathcal{IW}({\bm F}^j, {\mathcal P}^i, {\mathcal P}^j, \bm G) \in \mathbb{R}^{\frac{H}{4} \times \frac{W}{4} \times D},
\end{equation}
where $\mathcal{IW}$ represents the operation that only records the indices obtained during warping, and $G$ is the depth candidate matrix.
Next, we compute the feature correlations in parallel as follows:
\begin{equation}
\label{eq:correlation}
    {\bm C}^{i} = \mathbf{\Psi}({\bm F}^i, {\bm F}^j, {\bm I}^{j \to i}) \in \mathbb{R}^{\frac{H}{4} \times \frac{W}{4} \times D},
\end{equation}
where $\mathbf{\Psi}$ denotes the Sparse Matrix Multiplication (SMM), which uses the warp index ${\bm I}^{j \to i}$ to determine the position in ${\bm F}^j$ for matrix multiplication with ${\bm F}^i$.
Then, we refine the correlation map $\bm{C}^{i}$ with a lightweight 2D U-Net to obtain $\tilde{\bm{C}}^{i} \in \mathbb{R}^{H \times W \times D}$. Finally, a $\mathrm{softmax}$ is applied along the depth dimension to obtain the attention weights:

\begin{equation}
\bm{A}^{i}=\mathrm{softmax} (\tilde{\bm{C}}^{i}).
\end{equation}
This attention map $A^{i}$ corresponds to the depth probability results of view $i$ for different depth candidates in a single estimation.

\noindent \textbf{Depth Probability Boosting Strategy.}
In each depth probability boosting unit, we stack $M$ Warp-Index Epipolar Attention layers to produce $M$ depth probability estimation results.
To combine these isolated estimated outputs for stronger depth estimation capability, we propose a depth probability boosting strategy.
Specifically, we initialize the depth probability matrix $\bm P_{0}$ as an all-ones matrix and compute the subsequent updates as follows:
\begin{align}
\label{eq:hd_pa}
    \bm P_{m}= Norm(\bm P_{m-1} \odot \bm A_{m}),
\end{align}
where $m \in \{1,\ldots,M\}$ and $\mathrm{Norm}(\cdot)$ denotes row-wise normalization. $\bm{P}_{m}$ represents the depth probability matrix generated by the $m$-th Warp-Index Epipolar Attention in the current depth probability boosting unit.
Depth candidates with consistently high probabilities across layers will be boosted through this cascaded element-wise product process.
As a result, the depth probability produced by the index-based epipolar attention layer can be gradually enhanced, becoming more reliable and accurate.

\noindent \textbf{Iterative Depth Estimation Process.}
Based on the depth probability boosting strategy, each Depth Unit produces an enhanced depth-probability map $\bm{P}_{M,n}$ at iteration $n$.
To refine depth estimates symmetrically around the current prediction, we formulate the depth update using a relative depth-candidate offset vector $\Delta\bm{G}_{n}$, which allows the network to predict both positive and negative residuals.
Specifically, for the first iteration ($n=1$), we uniformly sample depth candidates within the initial range $[d_{\min}, d_{\max}]$. The depth-candidate vector used for feature matching is defined as $\bm{G}_{1} = [d_{\min}+{I}_{1},\ d_{\min}+2{I}_{1},\ \ldots,\ d_{\min}+\mathcal{D}_{1}{I}_{1}]$, where ${I}_{1} = (d_{\max}-d_{\min})/(\mathcal{D}_{1}+1)$ denotes the sampling interval and $\mathcal{D}_{1}$ is the number of candidates. Since the initial depth map is set to $\bm{D}_{0}=\bm{0}$, the relative offset vector in the first iteration is defined as $\Delta\bm{G}_{1}=\bm{G}_{1}$.
For subsequent iterations ($n>1$), the depth search range is centered at the previous depth estimate $\bm{D}_{n-1}$. Accordingly, the depth-candidate vector is updated as $\bm{G}_{n} = [\bm{D}_{n-1}-k{I}_{n},\ \ldots,\ \bm{D}_{n-1},\ \ldots,\ \bm{D}_{n-1}+k{I}_{n}]$, and the corresponding relative offset vector is $\Delta\bm{G}_{n} = [-k{I}_{n},\ \ldots,\ 0,\ \ldots,\ k{I}_{n}]$. Here, the interval shrinks as ${I}_{n} = {I}_{1}/n$, enabling progressively finer refinement over iterations, and the number of candidates is $\mathcal{D}_{n}=2k+1$.
The residual depth map $\Delta\bm{D}_{n}$ is then computed as the weighted sum of the relative offsets $\Delta\bm{G}_{n}$ with the probability map $\bm{P}_{M,n}$ along the depth dimension:
\begin{equation}
\label{eq:softmax_depth}
    {\Delta \bm D_{n}} = {\bm P_{M,n}} {\Delta \bm G_{n}}.
\end{equation}
The refined depth map is updated additively at each iteration:
\begin{equation}
\label{eq:depth_update}
    {\bm D_{n}} = {\bm D_{n-1}} + {\Delta \bm D_{n}},
\end{equation}
where $n \in \{ 1, \ldots, N \}$. ${\bm D_{N}}$ is the final depth map of the iterative depth estimation process, from which the Gaussian mean parameters are obtained through unprojection.
Enabled by our memory-efficient Warp-Index Epipolar Attention, IDESplat progressively increases the feature resolution over iterations.
In the final stage, IDESplat performs warping and similarity computation at the original input resolution of $256 \times 256$ to produce the refined depth map.

\subsection{Gaussian Focused Module}
\label{GFM}
For the remaining Gaussian parameters, we introduce a window-based Gaussian Focused Module that can filter irrelevant Gaussians and retains the most relevant tokens for attention weight computation.
The window-based attention performs pairwise interactions for each Gaussian token, often including numerous irrelevant ones.
This dense interaction not only slows down the model but also introduces noise into the attention results.
Inspired by \cite{long2025progressive}, we introduce a Gaussian Focused Layer that reuses the previous layer's Gaussian correlation map to guide attention in the current layer.
Formally, for the Gaussian parameters of a given view, $G \in \mathbb{R}^{C \times H \times W}$, we apply three linear layers to obtain $\mathbf{Q}$, $\mathbf{K}$, and $\mathbf{V}$.
We use a matrix $\mathbf{I}_{G}$ to record the indices of Gaussian tokens with high similarity.
$\mathbf{I}_{G}^{0}$ is initialized as an all-ones matrix and is updated after each similarity computation.
Then, we compute the Gaussian similarity map as follows:
\begin{equation} 
    \label{eq:sparse_sa_calculation}
    \mathbf{S}^{l} = \mathbf{\Psi} (\mathbf{Q}^{l}, \mathbf{K}^{l}, \mathbf{I}^{l-1}),
\end{equation}
where $\mathbf{\Psi}$ denotes the SMM operation, and $S^{l}$ is the similarity map of the $l$-th Gaussian-Focused Layer.
Subsequently, we compute the sparse Gaussian attention map $\mathbf{A}^{l}$ for the current $l$-th layer as follows:
\begin{align}
\label{eq:pfa_b}
    \mathbf{A}^{l} &= \mathcal{S}\!\left(\operatorname{Norm} \big( \mathbf{A}^{l-1} \odot Softmax(\mathbf{S}^{l})  \big)\right),
\end{align}
where $\mathcal{S}$ denotes the sparsification operation that retains the top half of the weights in each row of the attention map. The positions of the retained weights are recorded as $\mathbf{I}_{G}^{l}$, which stores the selected highly similar Gaussian relations.
Finally, the output Gaussian features are reweighted as follows:
\begin{equation}
\label{eq:attn_v}
    \mathbf{O}^{l} = \mathbf{\Psi}(\mathbf{A}^{l}, \mathbf{V}^{l}, \mathbf{I}^{l}).
\end{equation}
The Gaussian Focused Module consists of a series of Gaussian Focused Layers connected in sequence.
As the layer index $l$ increases, $\mathbf{I}_{G}^{l}$ becomes progressively sparser, gradually identifying the Gaussian positions that are most important to each query location.
This module leverages token similarity across layers to filter out the influence of irrelevant Gaussian features, achieving relational and sufficiently enriched Gaussian feature interactions.

\section{Experiments}
\label{sec:Experiments}

\subsection{Experiment Setting}

\noindent \textbf{Datasets.}~Our experiments are conducted on three large-scale datasets: 
RealEstate10K~\citep{DBLP:journals/tog/ZhouTFFS18}, ACID~\citep{liu2021infinite} and DTU~\citep{jensen2014large}. RealEstate10K contains real estate videos downloaded from YouTube, which are split into 67,477 training scenes and 7,289 testing scenes. ACID consists of nature scenes captured via aerial drones, with 11,075 scenes for training and 1,972 scenes for testing. Both datasets are calibrated using the Structure-from-Motion (SfM) ~\cite{schonberger2016structure} algorithm to estimate the camera's intrinsic and extrinsic parameters for each frame. Following the settings of previous works~\cite{charatan2024pixelsplat, chen2024mvsplat, liu2025monosplat, xu2025depthsplat}, for the RealEstate10K and ACID datasets, two context images are used as input, and three novel target views are rendered for each test scene. All input and target images have a resolution of $256\times256$. In addition, for the multi-view DTU~\cite{jensen2014large} dataset, which contains object-centric scenes with known camera poses, we report results on 16 validation scenes.

\noindent\textbf{Implementation details.}
For a fair comparison, we followed the commonly used training setup~\cite{chen2024mvsplat, liu2025monosplat, xu2025depthsplat}. Specifically, our training experiments were conducted on 8 RTX 4090 GPUs with a total batch size of 16, using the AdamW~\citep{loshchilov2017decoupled} optimizer for 300,000 iterations. We employed a cosine learning rate schedule to optimize our model. For the pre-trained Depth Anything V2~\cite{yang2024depth} backbone, we used a learning rate of $2 \times 10^{-6}$, while the remaining layers were trained with a learning rate of $2 \times 10^{-4}$. Following DepthSplat~\cite{xu2025depthsplat}, we train our model with MSE and LPIPS losses.

%%%% 添加更多细节

\subsection{Main Results}
We comprehensively compare our proposed method with several leading approaches in scene-level novel view synthesis, covering three representative categories: light field network methods such as GPNR~\citep{suhail2022generalizable} and AttnRend~\citep{du2023learning}; NeRF-based methods including pixelNeRF~\citep{yu2021pixelnerf} and MuRF~\cite{xu2024murf}; and 3D Gaussian Splatting-based methods such as pixelSplat~\citep{charatan2024pixelsplat}, latentSplat~\cite{wewer2024latentsplat}, MVSplat~\citep{chen2024mvsplat}, eFreeSplat~\cite{min2024epipolar}, MonoSplat~\cite{liu2025monosplat}, and DepthSplat~\cite{xu2025depthsplat}.

\begin{table*}[t]
% \captionsetup{font={small}}
% \scriptsize
\normalsize
\centering
% \footnotesize

\caption{
\textbf{Quantitative comparisons.} We surpass all baseline methods in terms of PSNR, LPIPS, and SSIM for novel view synthesis on the real-world RealEstate10k~\cite{DBLP:journals/tog/ZhouTFFS18} and ACID~\cite{liu2021infinite} datasets. We highlight first-place results in \textbf{bold} and second-place results with \underline{underlines} in each column. "-" Indicates that the original paper did not contain relevant data.}
% \begin{tabular}{p{3.8cm}|c|ccc|ccc}
\begin{tabular}{l|c|ccc|ccc}
\toprule

\multirow{2}{*}{\textbf{Method}} & \multirow{2}{*}{\textbf{Params (M) $\downarrow$}} & \multicolumn{3}{c|}{\textbf{RealEstate10k~\cite{DBLP:journals/tog/ZhouTFFS18}}} & \multicolumn{3}{c}{\textbf{ACID~\cite{liu2021infinite}}}    \\

&   &  PSNR $\uparrow$ & SSIM $\uparrow$ & LPIPS $\downarrow$ & PSNR $\uparrow$ & SSIM $\uparrow$ & LPIPS $\downarrow$\\

\midrule
Du et al.\cite{du2023learning}            & 125.1           & 24.78 & 0.820 & 0.213 & 26.88 & 0.799 & 0.218  \\
GPNR\cite{suhail2022generalizable}        & \underline{9.6} & 24.11 & 0.793 & 0.255 & 25.28 & 0.764 & 0.332  \\
pixelNeRF~\cite{yu2021pixelnerf}          & 28.2            & 20.43 & 0.589 & 0.550 & 20.97 & 0.547 & 0.533  \\
MuRF~\cite{xu2024murf}                    & \textbf{5.3}    & 26.10 & 0.858 & 0.143 & 28.09 & 0.841 & 0.155  \\
\midrule
pixelSplat~\cite{charatan2024pixelsplat}~(CVPR 2024)  & 125.4           & 26.09 & 0.863 & 0.136 & 28.27 & 0.843 & 0.146   \\
latentSplat~\cite{wewer2024latentsplat}~(ECCV 2024)   & 187.0           & 23.07 & 0.825 & 0.182 & 24.95 & 0.782 & 0.207   \\
MVSplat~\cite{chen2024mvsplat}~(ECCV 2024)            & 12.0            & 26.39 & {0.869} & 0.128 & 28.25 & 0.843 & 0.144  \\
eFreeSplat~\cite{min2024epipolar}~(NIPS 2024)         & -             & 26.45 & 0.865 & {0.126} & {28.30} & {0.851} & {0.140}  \\
MonoSplat~\cite{liu2025monosplat}~(CVPR 2025)         & 30.3            & {26.68} & {0.875} & {0.123} & \underline{28.63} & \underline{0.864} & \underline{0.138}  \\
DepthSplat~\cite{xu2025depthsplat}~(CVPR 2025)        & 354             & \underline{27.47} & \underline{0.889} & \underline{0.114} & {-} & {-} & {-}  \\
\rowcolor{row_color}
\textbf{IDESplat (Ours)}                  & 37.6            & \textbf{27.80} & \textbf{0.893} & \textbf{0.108} & \textbf{28.94} & \textbf{0.866} & \textbf{0.130}  \\
\bottomrule
\end{tabular}

\label{tab:comparison_sota}
\end{table*}

\noindent\textbf{Quantitative results.}
As shown in Tab.~\ref{tab:comparison_sota}, our IDESplat achieves state-of-the-art performance on all visual quality metrics in both the RealEstate10K and ACID benchmarks.
Specifically, on the RE10K dataset, compared to DepthSplat, our method improves PSNR by \textbf{0.33 dB}, while using only \textbf{10.7\%} of its parameters. 
Additionally, compared to MonoSplat, which has a similar number of parameters, our method achieves a significant \textbf{1.12 dB} improvement.
For SSIM and LPIPS metrics, our method also achieves the best results of 0.893 and 0.108, respectively.
On the ACID dataset, our method achieves a \textbf{0.31 dB} improvement in PSNR over MonoSplat, reaching a maximum of \textbf{28.94 dB}.
The superior performance of IDESplat can be attributed to its iterative boosting design for depth probability prediction, which gradually refines the depth map.
This design allows our method to achieve reliable and stable depth estimation results with fewer parameters, enabling accurate Gaussian mean parameter prediction and high-quality scene reconstruction.

\noindent\textbf{Cross-Dataset Generalization.}
To evaluate the generalization ability of our proposed IDESplat, we conducted cross-dataset generalization tests on unseen datasets.
We first trained the model on the indoor scene dataset RealEstate10K, and then evaluated it directly on the outdoor scene ACID dataset and the object-centered DTU dataset.
As shown in Tab.~\ref{tab:cross_dataset}, IDESplat demonstrates outstanding cross-dataset generalization ability, outperforming existing methods on all metrics.
For the DTU dataset, our method improves PSNR by \textbf{2.95 dB} over DepthSplat, and achieves the best LPIPS score of \textbf{0.239}. On the ACID dataset, our method also achieves the highest performance with \textbf{28.79 dB}.
These results show that our iterative depth probability estimation method effectively enhances the modeling of cross-view feature similarity, learning strong out-of-distribution scene reconstruction capabilities.
This also demonstrates that our iterative depth probability estimation method is both efficient and highly generalizable, surpassing dataset-specific features and avoiding overfitting to the training data.

\begin{figure*}
    \centering
    \includegraphics[width=\linewidth]{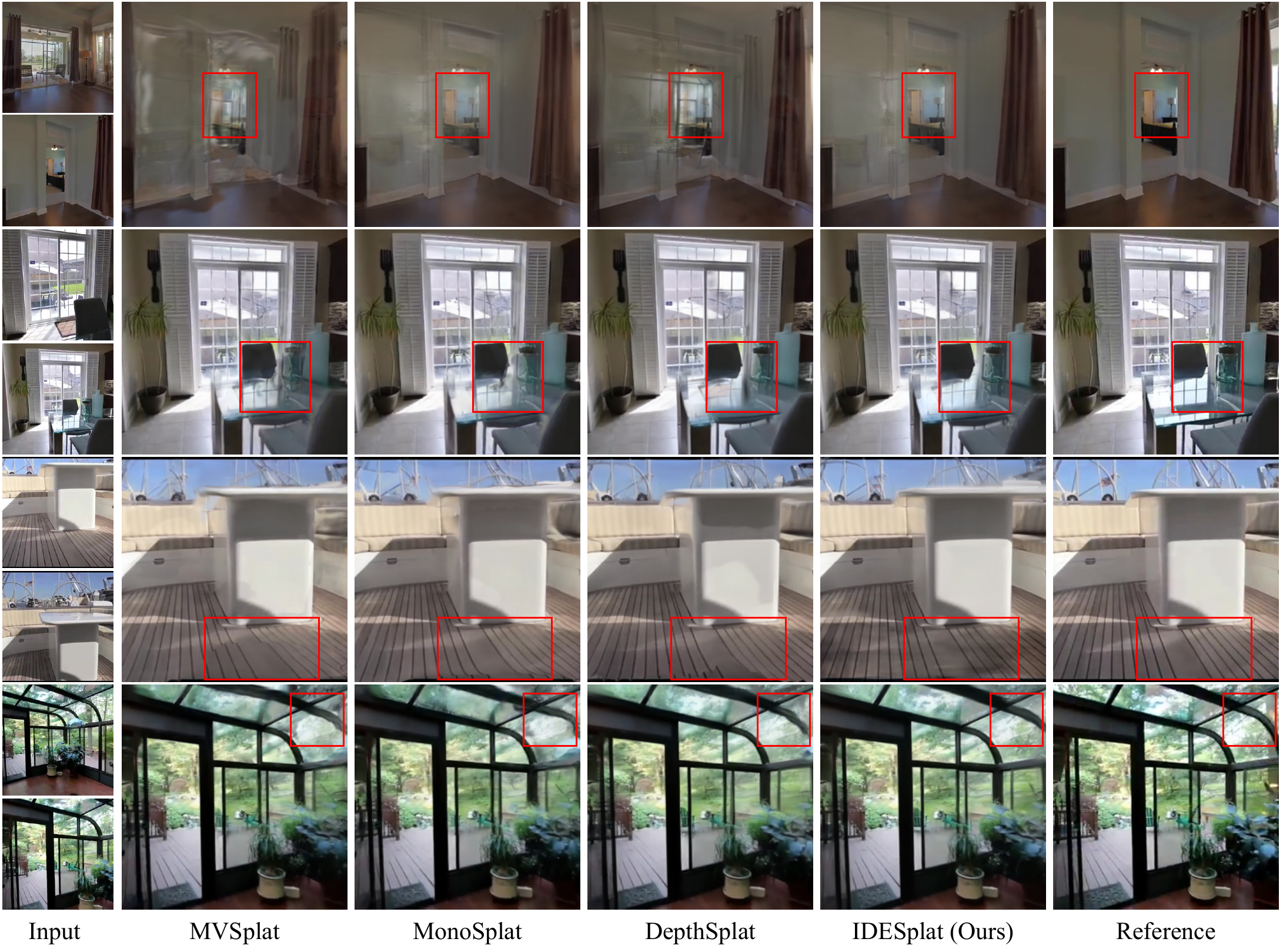}
    \caption{The comparison of visualization results for novel view synthesis on the RealEstate10K dataset. Our IDESplat significantly outperforms previous state-of-the-art methods in rendering challenging regions.}
    \label{fig:visual_nvs}
    \vspace{-3mm}
\end{figure*}

\noindent\textbf{Visual comparison results.}
We performed a qualitative comparison of scene reconstruction results between our method and mainstream models, including MVSplat~\cite{chen2024mvsplat}, MonoSplat~\cite{liu2025monosplat}, and DepthSplat~\cite{xu2025depthsplat}. The visual comparison results are shown in Fig.~\ref{fig:visual_nvs}, which include both indoor and outdoor scenes.
Our proposed IDESplat achieves significantly better novel view synthesis results compared to existing mainstream open-source models.
Specifically, in challenging areas with rich textures and large position and angle differences between the input images, existing methods often produce noticeable artifacts and blurring in the synthesized views.
In contrast, IDESplat is able to generate high-quality reconstructions in these difficult regions, demonstrating excellent texture consistency and clearly showing its superiority in reconstructing complex scenes.

\begin{table}
    \footnotesize
    \centering
    \caption{\textbf{Quantative comparisons of cross-dataset generalization.} We perform zero-shot tests on ACID~\cite{liu2021infinite} and DTU~\cite{jensen2014large} datasets, using models trained solely on RealEstate10K~\cite{DBLP:journals/tog/ZhouTFFS18}. Best and second best results are \textbf{bolded} and \underline{underlined}.}
    \vspace{-2mm}
    % Reduce column separation
    \setlength{\tabcolsep}{2.5pt} % Adjust this value as needed
    \begin{tabular}{l|ccc|ccc}
        \toprule
        \multirow{2}{*}{\textbf{Method}} & \multicolumn{3}{c|}{\textbf{RE10k$\rightarrow$DTU}} & \multicolumn{3}{c}{\textbf{RE10k$\rightarrow$ACID}}\\
        & PSNR$\uparrow$ & SSIM$\uparrow$ & LPIPS$\downarrow$  & PSNR$\uparrow$ & SSIM$\uparrow$ & LPIPS$\downarrow$ \\
        \midrule
        pixelSplat~\cite{charatan2024pixelsplat} & 12.89             & 0.382             & 0.560 & 27.64 & 0.830 & 0.160 \\
        MVSplat~\cite{chen2024mvsplat}           & 13.94             & 0.473             & 0.385 & 28.15 & 0.841 & 0.147 \\
        MonoSplat~\cite{liu2025monosplat}        & 15.25             & \underline{0.605} & \underline{0.291} & 28.24 & \underline{0.848} & 0.145 \\
        Depthsplat~\cite{xu2025depthsplat}       & \underline{15.38} & 0.415             & 0.442 & \underline{28.37} & 0.847 & \underline{0.141} \\
        \rowcolor{row_color}
        \textbf{IDESplat (Ours)}                 & \textbf{18.33}    & \textbf{0.719}    & \textbf{0.239} & \textbf{28.79} & \textbf{0.853} & \textbf{0.135} \\
        \bottomrule
    \end{tabular}
    \label{tab:cross_dataset}
\vspace{-6mm}
\end{table}

\begin{figure*}
    \centering
    \includegraphics[width=\linewidth]{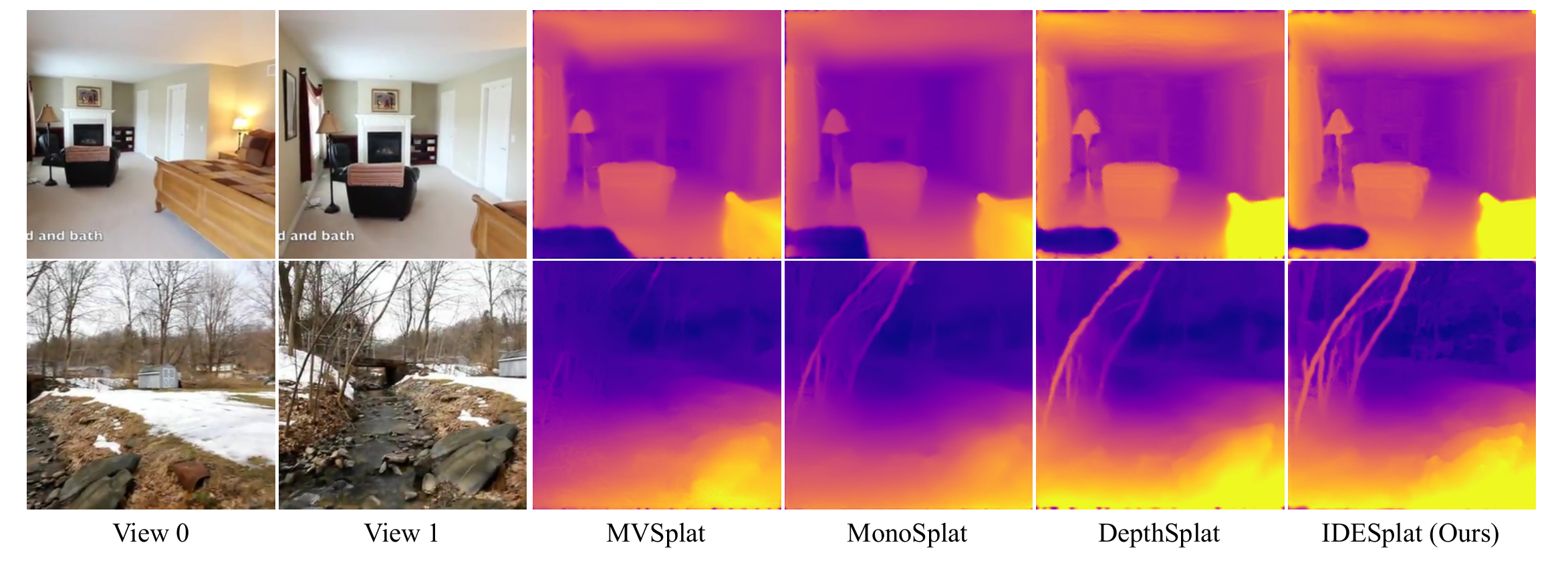}
    \vspace{-8mm}
    \caption{Comparison of depth prediction maps for different models on the RE10K dataset.}
    \label{fig:visual_depth}
    \vspace{-6mm}
\end{figure*}

\noindent\textbf{Visual comparison of depth maps.}
We conducted a direct qualitative comparison of depth maps predicted by IDESplat and existing state-of-the-art models, including MVSplat~\cite{chen2024mvsplat}, MonoSplat~\cite{liu2025monosplat}, and DepthSplat~\cite{xu2025depthsplat}. It is important to note that the generalizable 3DGS task typically does not have depth ground truth, so we analyzed the results using the corresponding reference RGB images as input. 
As shown in Fig.~\ref{fig:visual_depth}, IDESplat consistently outperforms existing methods in both indoor and outdoor scenes.
Even in regions with similar foreground and background appearances or rich texture details, it can better distinguish objects at different depths and produce more realistic depth estimates with finer details.
This advantage comes from aggregating feature similarity across multiple warps via the depth probability boosting strategy, as well as performing warp computation at the original scale to better preserve fine textures. 
More visual results are provided in the supplementary materials.

\noindent\textbf{Comparison of model efficiency.}
In Tab.~\ref{tab:model_efficiency}, we compare the model efficiency of IDESplat with several classic methods, including PixelSplat, MVSplat, and the latest methods such as MonoSplat and DepthSplat. We measure inference time and memory usage with two-view inputs at a resolution of $256 \times 256$, and the PSNR values are reported on the commonly used benchmark dataset RE10K. Although our method shows slightly lower inference efficiency compared to DepthSplat, it outperforms DepthSplat on all other metrics. Despite using significantly fewer parameters and computational resources than DepthSplat, IDESplat achieves better results. The superior performance of IDESplat with fewer parameters and better memory efficiency is due to our iterative depth probability estimation architecture, which performs multiple warp operations to improve the accuracy of Gaussian mean parameter predictions.

\begin{table}
    \footnotesize
    \centering
    \caption{\textbf{Comparison of model efficiency.}  Our method demonstrates relatively low inference costs and reduced memory usage. }
    \vspace{-2mm}
    % Reduce column separation
    \setlength{\tabcolsep}{5pt} % Adjust this value as needed
    \begin{tabular}{c|ccc|c}
        \toprule
        Method                                   & Params (M)   & Mem. (M)  & Time (s)  &  PSNR$\uparrow$ \\
        \midrule
        pixelSplat~\cite{charatan2024pixelsplat} & 125.4        & 4108      & 0.120     & 26.09 \\
        MVSplat~\cite{chen2024mvsplat}           & \textbf{12.0}& 1940      & \textbf{0.054}     & 26.39 \\
        MonoSplat~\cite{liu2025monosplat}        & 30.3         & \textbf{1606}& 0.062     & 26.68 \\
        Depthsplat~\cite{xu2025depthsplat}       & 354          & 3342      & 0.082     & 27.47 \\
        \rowcolor{row_color}
        \textbf{IDESplat (Ours)}                 & 37.6         & 2336      & 0.110     & \textbf{27.80} \\

        \bottomrule
    \end{tabular}
        \label{tab:model_efficiency}
        \vspace{-2mm}
\end{table}

\begin{figure}[htp]
    \centering
    % \vspace{-2mm}
    \includegraphics[width=\linewidth]{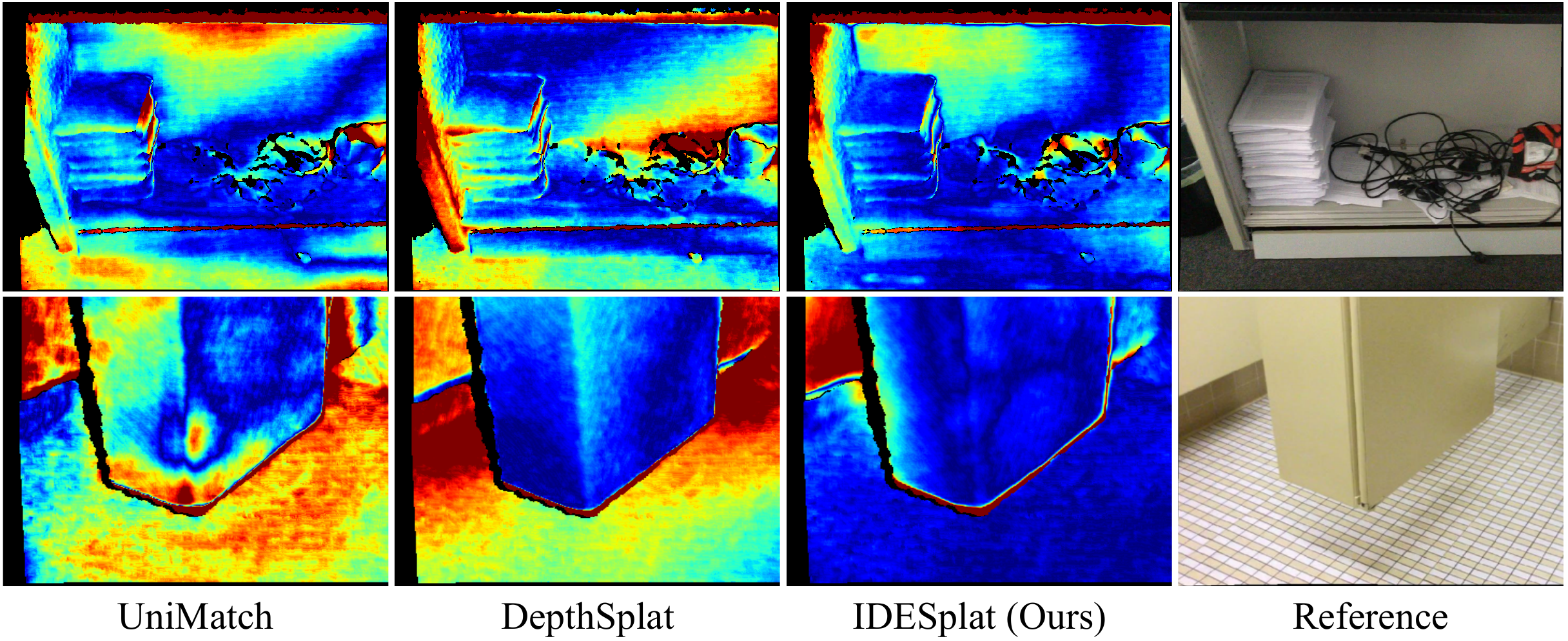}
    \vspace{-6mm}
    \caption{Qualitative comparison of depth error maps on the ScanNet dataset.}
    \vspace{-6mm}
    \label{fig:scannet_depth_error_map}
\end{figure}

\noindent\textbf{Depth estimation experiment on ScanNet.}
We report depth estimation results on the ScanNet dataset in Table~\ref{tab:scannet_depth}. IDESplat consistently outperforms UniMatch and DepthSplat across standard quantitative depth estimation metrics.
For qualitative comparison, we further visualize the depth error maps on the ScanNet dataset in Fig.~\ref{fig:scannet_depth_error_map}. Compared with the baseline methods, IDESplat exhibits fewer large-error regions and produces more accurate depth estimates, especially around geometrically complex areas.

\begin{table}[htbp]
\footnotesize
\centering
\vspace{-2mm}
% --- 在这里调整行距 ---
\renewcommand{\arraystretch}{1.0} % 默认是1.0，1.2表示拉大20%
% --------------------
\setlength{\tabcolsep}{15pt} % increase column spacing
\caption{\textbf{Depth estimation results on the ScanNet dataset.}}
\vspace{-2mm}
\resizebox{0.98\linewidth}{!}{
\begin{tabular}{l|ccc}
    \toprule
    Method & Abs Rel $\downarrow$ & RMSE $\downarrow$ & RMSE$_{\mathrm{log}}$ $\downarrow$ \\
    \midrule
    UniMatch   & 0.059 & 0.179 & 0.082 \\
    DepthSplat & 0.045 & 0.125 & 0.061 \\
    \rowcolor{row_color}
    IDESplat   & \textbf{0.039} & \textbf{0.116} & \textbf{0.053} \\
    \bottomrule
\end{tabular}}
\label{tab:scannet_depth}
\vspace{-4mm}
\end{table}

\subsection{Ablation Study}

We conducted detailed ablation experiments on various components of the proposed IDESplat. 
All models were trained for 20,000 iterations on the RealEstate10K dataset with a batch size of 8.
We analyzed the effectiveness of the Gaussian Focused Module (GFM), Iterative Depth Estimation process (IDE), and Depth Probability Boosting Strategy (DPBS) , while keeping all other training settings consistent.
It should be noted that IDE(3) indicates the model iterates the depth probability boosting unit three times.
The detailed experimental results, shown in Tab.~\ref{tab:ablation_module}, demonstrate that all three designs contribute to improved reconstruction performance. 
Specifically, adding GFM and IDE(3) results in improvements of 0.32 dB and 0.57 dB, respectively, showing that both the iterative depth estimation process and the Gaussian focused module effectively enhance scene reconstruction.
The performance improves significantly by 0.46 dB when we incorporate the DPBS strategy into the iterative process, and the best result of 27.56 dB is achieved when all three strategies are used together, further validating the effectiveness of our proposed method.

\begin{table}
\small
\centering
    \caption{\textbf{Ablation study results for each component of IDESplat.} All results are reported on the RealEstate10K dataset. GFM denotes the Gaussian Focused Module, IDE(3) denotes the Iterative Depth Estimation process with three iterations, and DPBS denotes the Depth Probability Boosting Strategy.}
    \vspace{-2mm}
    % Reduce column separation
    \setlength{\tabcolsep}{10pt} % Adjust this value as needed
    \begin{tabular}{l|ccc}
        \toprule
        Method & PSNR$\uparrow$ & SSIM$\uparrow$ & LPIPS$\downarrow$ \\
        \midrule
        Baseline            & 26.31   & 0.866   & 0.129  \\
        \midrule
         + GFM              & 26.63   & 0.875  & 0.124 \\
         + IDE(3)           & 26.88   & 0.878  & 0.120 \\
         + IDE(3) + GFM     & 27.07   & 0.882  & 0.118 \\
         + IDE(3) + DPBS     & 27.34   & 0.887  & 0.112 \\
         Full Model         & \textbf{27.56}   & \textbf{0.889}  & \textbf{0.110} \\
        \bottomrule
    \end{tabular}
    \label{tab:ablation_module}
    \vspace{-6mm}
\end{table}

We conduct ablation experiments to evaluate the efficiency and reconstruction performance of IDESplat under different iteration counts.
In these experiments, the Gaussian Focused Module and Depth Probability Boosting Strategy were used by default.
Additionally, during the iterative process, we gradually increased the resolution, and the depth search range was progressively halved. For 3 iterations, the feature resolutions during warping were $\frac{1}{4}$, $\frac{1}{2}$, and $1$ of the original resolution.
For 4 iterations, the model iterates twice at the largest feature size. When the number of iterations is 0, it represents performing a single warp for depth probability estimation.
The results of the experiment, as shown in Tab.~\ref{tab:ablation_iterations}, indicate that even with a single iteration, our method provides a 0.45 dB improvement.
With 3 iterations, the performance improves by 0.93 dB. Increasing the iteration count adds little parameter overhead and keeps memory usage comparable to existing methods. Despite the longer inference time, 3 iterations already provide a substantial gain while maintaining real-time inference.

\begin{table}
    \footnotesize
    \centering
    \caption{\textbf{Ablation results for IDESplat with different numbers of iterative depth probability boosting units.}  All results are reported on the RealEstate10K dataset. Here, an iteration count of 0 denotes using a single warp operation without any DPBUs, while the other entries correspond to using different numbers of DPBUs.}
    \vspace{-2mm}
    % Reduce column separation
    \setlength{\tabcolsep}{2.2pt} % Adjust this value as needed
    \begin{tabular}{c|ccc|ccc}
        \toprule
        Iterations & Params (M) & Mem. (M) & Time (s) & PSNR$\uparrow$ & SSIM$\uparrow$ & LPIPS$\downarrow$ \\
        \midrule
        0           & 35.4 & 1674 & 0.056 & 26.63 & 0.875 & 0.124 \\
        1           & 36.8 & 1734 & 0.071 & 27.08 & 0.882 & 0.113\\
        2           & 37.3 & 1902 & 0.091 & 27.31 & 0.884 & 0.112\\
        3           & 37.6 & 2336 & 0.110 & 27.56 & 0.889 & 0.110\\
        4           & 38.0 & 2745 & 0.132 & 27.64 & 0.890 & 0.109\\
        \bottomrule
    \end{tabular}
        \label{tab:ablation_iterations}
        \vspace{-8mm}
\end{table}

\section{Conclusion}
\label{sec:conclusion}
We propose IDESplat, which iteratively applies warp operations and integrates multi-level epipolar attention maps to enhance depth probability estimation for accurate Gaussian mean prediction.
Specifically, we first design a depth probability boosting unit to amplify cross-view feature similarity in a multiplicative manner.
Then, we build an iterative depth estimation process by stacking multiple DPBUs, gradually identifying the most likely depth locations.
Additionally, during this iterative process, we progressively narrow the depth search range and increase feature size to achieve more precise depth estimates.
For the other Gaussian parameters, we design a gaussian focused module to select the most relevant Gaussian tokens for attention-based feature interaction.
Experimental results on large-scale benchmarks clearly show that IDESplat achieves excellent reconstruction quality and strong domain generalization ability.
%
% The superior performance of IDESplat can be attributed to its iteratively enhanced depth probability prediction design, which progressively refines the depth map.
%
Our current model still has limitations: although it enables real-time 3D reconstruction, its efficiency could be further improved, and it requires camera pose input as a prerequisite.
Potential directions for improvement include further enhancing inference speed and exploring a pose-free framework.

\section*{Acknowledgement}
% \vspace{2mm}
% This work was supported by the National Natural Science Foundation of China (No.~62476051).
{\raggedright
This work was supported by the National Natural Science Foundation of China under Grant No.~62476051.\par
}

{
    \small
    \bibliographystyle{ieeenat_fullname}
    \bibliography{main}
}

% WARNING: do not forget to delete the supplementary pages from your submission 
\clearpage
\setcounter{page}{1}
\maketitlesupplementary

In this supplementary material, we provide additional details on model training, model architecture, ablation study on DPBU, experimental results on the DL3DV dataset, and more visual comparison results. Specifically, in Section A, we present the training details of the IDESplat model. In Section B, we provide the model architecture details of IDESplat and Warp-Index Epipolar Attention. In Section C, we present the experimental results of IDESplat on the DL3DV dataset, along with the ablation study on Depth Probability Boosting Units. Finally, in Section D, we include more visual comparison results for novel view synthesis and depth prediction.

% \vspace{-2mm}
\section*{A. Training Details}
\label{sec:training_details}

For a fair comparison, we trained our IDESplat using a standard setup~\cite{chen2024mvsplat, liu2025monosplat, xu2025depthsplat}. The training was done on 8 RTX 4090 GPUs with a batch size of 16, using the AdamW~\citep{loshchilov2017decoupled} optimizer for 300,000 iterations, which took approximately 3 days. We used a cosine learning rate schedule. For the pre-trained Depth Anything V2~\cite{yang2024depth} backbone, the learning rate was $2 \times 10^{-6}$, while other layers were trained with a learning rate of $2 \times 10^{-4}$. The network was trained with a combination of MSE and LPIPS losses between the rendered and ground truth images. Following~\cite{xu2025depthsplat}, for the newly added DL3DV dataset, we trained at a resolution of $256 \times 448$. First, we pre-trained on RE10k and then fine-tuned on the DL3DV dataset for 100K iterations, with a total batch size of 4, and the number of input views was randomly sampled from 2 to 6. During inference, we evaluated the model’s performance on different numbers of input views.

% \vspace{-2mm}
\section*{B. Model Architecture Details}
\label{sec:model_arch_details}

We provide a detailed description of the IDESplat network architecture, as shown in Figure~\ref{fig:IDESplat}. It consists of three main parts: a feature extraction backbone, an iterative depth estimation process, and a Gaussian focus module. The backbone has two branches: a multi-view branch using the pre-trained Unimatch~\citep{xu2023unifying} and a monocular branch using the ViT-small version of DepthAnything V2~\cite{yang2024depth}. The outputs from both branches are fused to provide multi-view geometry and texture information for the next modules.
The depth estimation process includes three Depth Probability Boosting Units (DPBU) that sequentially generate optimized depth results. Each DPBU contains two cascaded Warp-Index Epipolar Attention layers, which use the Hadamard product to enhance depth probabilities. The process is repeated for six transformations at resolutions of $64 \times 64$, $128 \times 128$, and $256 \times 256$.
The GFM has six layers, using a shifting window strategy with a window size of 16. After each attention calculation, the top half of the most relevant Gaussian positions are retained. The number of retained Gaussian weights per layer is [256, 256, 128, 128, 64, 64], and the module uses 6 attention heads with 256 channels in total.

To address the memory issues in existing warp computations for cross-view similarity, we introduce Warp-Index Epipolar Attention. Unlike the existing method, which samples target view features for each depth candidate and consumes a lot of memory, our approach only stores transformation indices for similarity matrix multiplication and uses Sparse Matrix Multiplication (SMM) for efficient computation. This design enables IDESplat to perform multiple rounds of warp and depth estimation more efficiently.

\begin{figure*}[htbp]
    \centering
    \includegraphics[width=\linewidth]{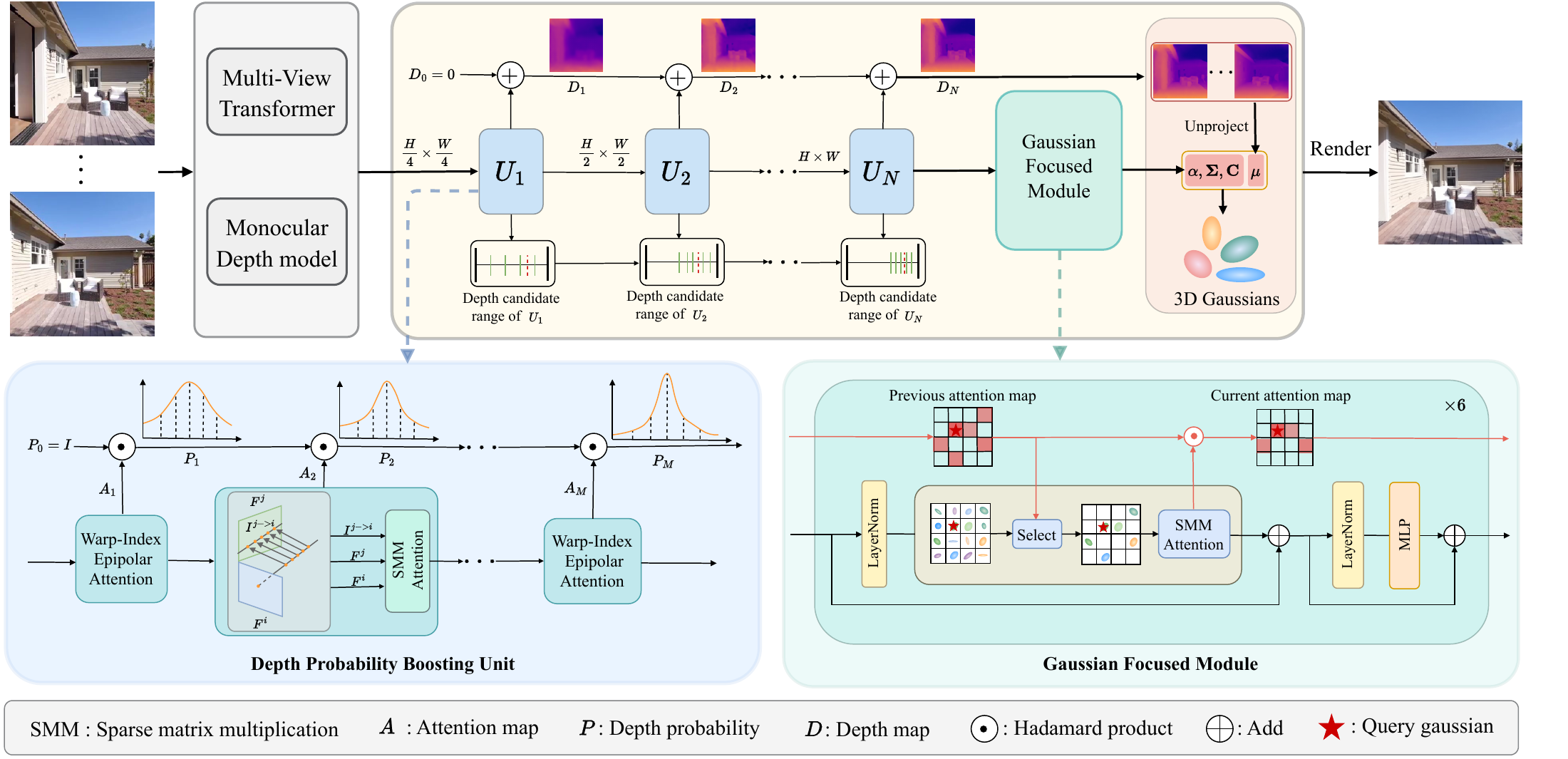}
    \caption{The architecture of IDESplat. IDESplat comprises a feature extraction backbone, an iterative depth estimation process with cascaded Depth Probability Boosting Units (DPBUs), and a Gaussian Focused Module (GFM).}
    \label{fig:IDESplat}
    \vspace{-3mm}
\end{figure*}

\begin{figure}[htbp]
    \centering
    \includegraphics[width=\linewidth]{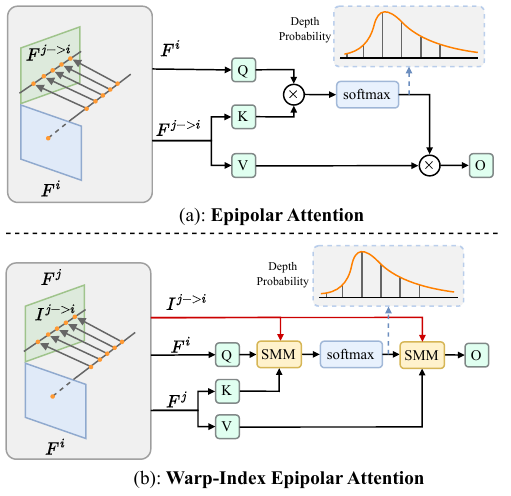}
    \vspace{-4mm}
    \caption{The difference between Warp-Index Epipolar Attention and Epipolar Attention.}
    \label{fig:WIEA}
    \vspace{-6mm}
\end{figure}

\section*{C. More Experimental Results}
\label{sec:more_exp_results}

We conducted additional experiments on the DL3DV dataset to further evaluate the proposed IDESplat method. DL3DV is a large-scale real-world multi-view video dataset, which helps validate our method's reconstruction capability in more complex and larger scenes. The experimental results, shown in Table~\ref{tab:dl3dv_compare}, demonstrate outstanding performance of our method compared to existing MVSplat and DepthSplat methods. IDESplat outperforms DepthSplat by \textbf{0.62dB}, \textbf{0.41dB}, and\textbf{ 0.42dB} when using 2, 4, and 6 input views, respectively. These results clearly show that our IDESplat provides better reconstruction performance in large, complex scenes with multiple input views compared to existing methods. 

We also conducted ablation experiments on DPBU with different numbers of Warp-Index Epipolar Attention. The results in Table~\ref{tab:ablation_DPBUs} show that depth probabilities can only undergo Multiplicative Boosting when more than one Warp-Index Epipolar Attention is used. Our IDESplat performs well when the number of attention layers exceeds one, and the performance improves as the number of layers increases. Considering both efficiency and performance, we chose the model with two Warp-Index Epipolar Attention layers.

\begin{table}[t]

\begin{center}
\small

\caption{\textbf{Quantitative Experimental Results and Comparisons on DL3DV}. Our IDESplat consistently outperforms MVSplat and DepthSplat across different numbers of input views. 
}
\vspace{-2mm}
\resizebox{\linewidth}{!}{
\begin{tabular}{lcccccccccccccccccccccc}
\toprule
Method & \#Views & PSNR $\uparrow$ & SSIM $\uparrow$ & LPIPS $\downarrow$\\
    
\toprule

MVSplat~\cite{chen2024mvsplat} & \multirow{2}{*}[-2pt]{2} & 17.54 & 0.529 & 0.402\\
DepthSplat~\cite{xu2025depthsplat} & & 19.31 & 0.615 & 0.310 \\
IDESplat & & \textbf{19.93} & \textbf{0.635} & \textbf{0.300} \\

\midrule

MVSplat~\cite{chen2024mvsplat} & \multirow{2}{*}[-2pt]{4} & 21.63 & 0.721 & 0.233 \\
DepthSplat~\cite{xu2025depthsplat} & & 23.12 & 0.780 & 0.178\\
IDESplat & & \textbf{23.53} & \textbf{0.789} & \textbf{0.176} \\

\midrule

MVSplat~\cite{chen2024mvsplat} & \multirow{2}{*}[-2pt]{6} & 22.93 & 0.775 & 0.193 \\
DepthSplat~\cite{xu2025depthsplat} & & 24.19 & 0.823 & 0.147 \\
IDESplat & & \textbf{24.61} & \textbf{0.829} & \textbf{0.146} \\

\bottomrule
\end{tabular}
\label{tab:dl3dv_compare}
\vspace{-6mm}
}
\end{center}
\end{table}

\begin{table}[htbp]
    \footnotesize
    \vspace{-6mm}
    \centering
    \caption{\textbf{Ablation results for DPBU with different numbers of Warp-Index Epipolar Attention.}  All results are reported on the RealEstate10K dataset. 
    % Here, an iteration count of 0 denotes using a single warp operation without any DPBUs, while the other entries correspond to using different numbers of DPBUs.
    }
    \vspace{-2mm}
    % Reduce column separation
    \setlength{\tabcolsep}{1.2pt} % Adjust this value as needed
    \begin{tabular}{c|ccc|ccc}
        \toprule
        Number of WIEA & Params (M) & Mem. (M) & Time (s) & PSNR$\uparrow$ & SSIM$\uparrow$ & LPIPS$\downarrow$ \\
        \midrule
        1           & 36.4 & 2033 & 0.082 & 27.11 & 0.882 & 0.116 \\
        2           & 37.6 & 2336 & 0.110 & 27.56 & 0.889 & 0.110 \\
        3           & 38.9 & 2642 & 0.139 & 27.65 & 0.890 & 0.109 \\
        4           & 40.2 & 2954 & 0.172 & 27.72 & 0.893 & 0.107 \\
        \bottomrule
    \end{tabular}
        \label{tab:ablation_DPBUs}
        \vspace{-6mm}
\end{table}

\section*{D. More Visual Comparison Results}
\label{sec:more_exp_results}

\begin{figure*}
    \centering
    \includegraphics[width=\linewidth]{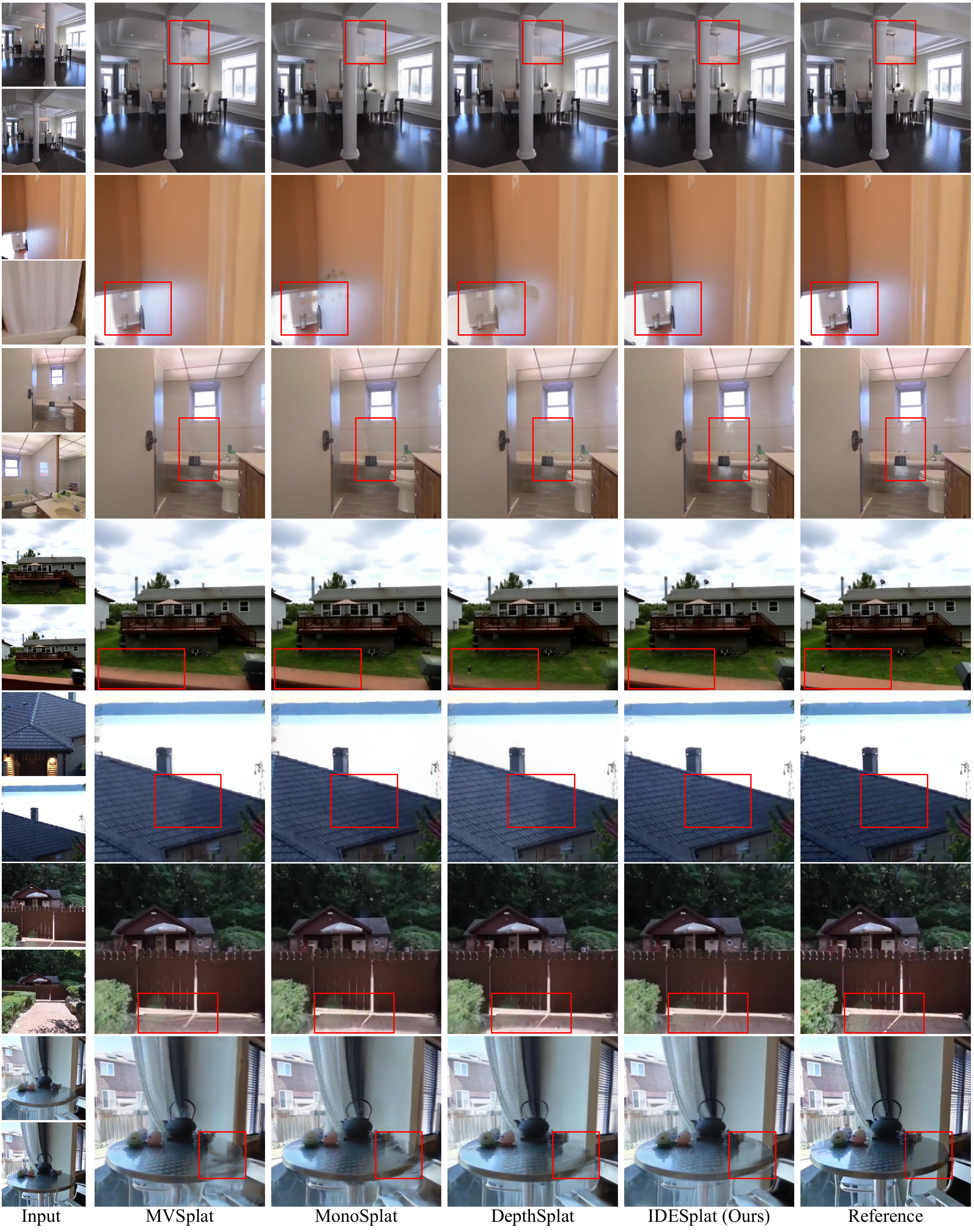}
    \caption{The comparison of visualization results for novel view synthesis on the RealEstate10K dataset. }
    \label{fig:suppl_re10k_visual}
    \vspace{-3mm}
\end{figure*}

\begin{figure*}
    \centering
    \includegraphics[width=\linewidth]{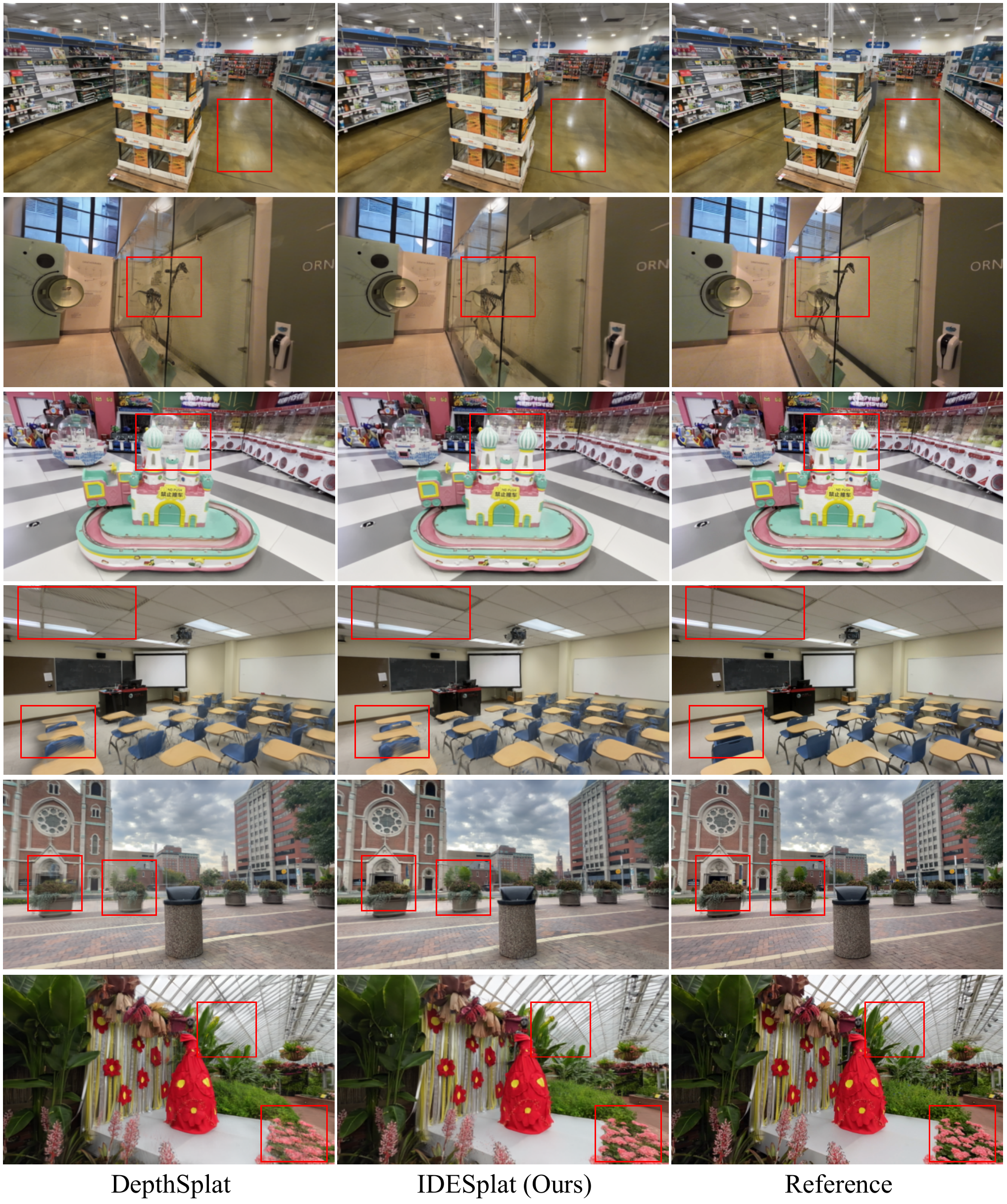}
    \caption{The comparison of visualization results for novel view synthesis on the DL3DV dataset.}
    \label{fig:visual_suppl_DL3DV}
    \vspace{-3mm}
\end{figure*}

\begin{figure*}
    \centering
    \includegraphics[width=\linewidth]{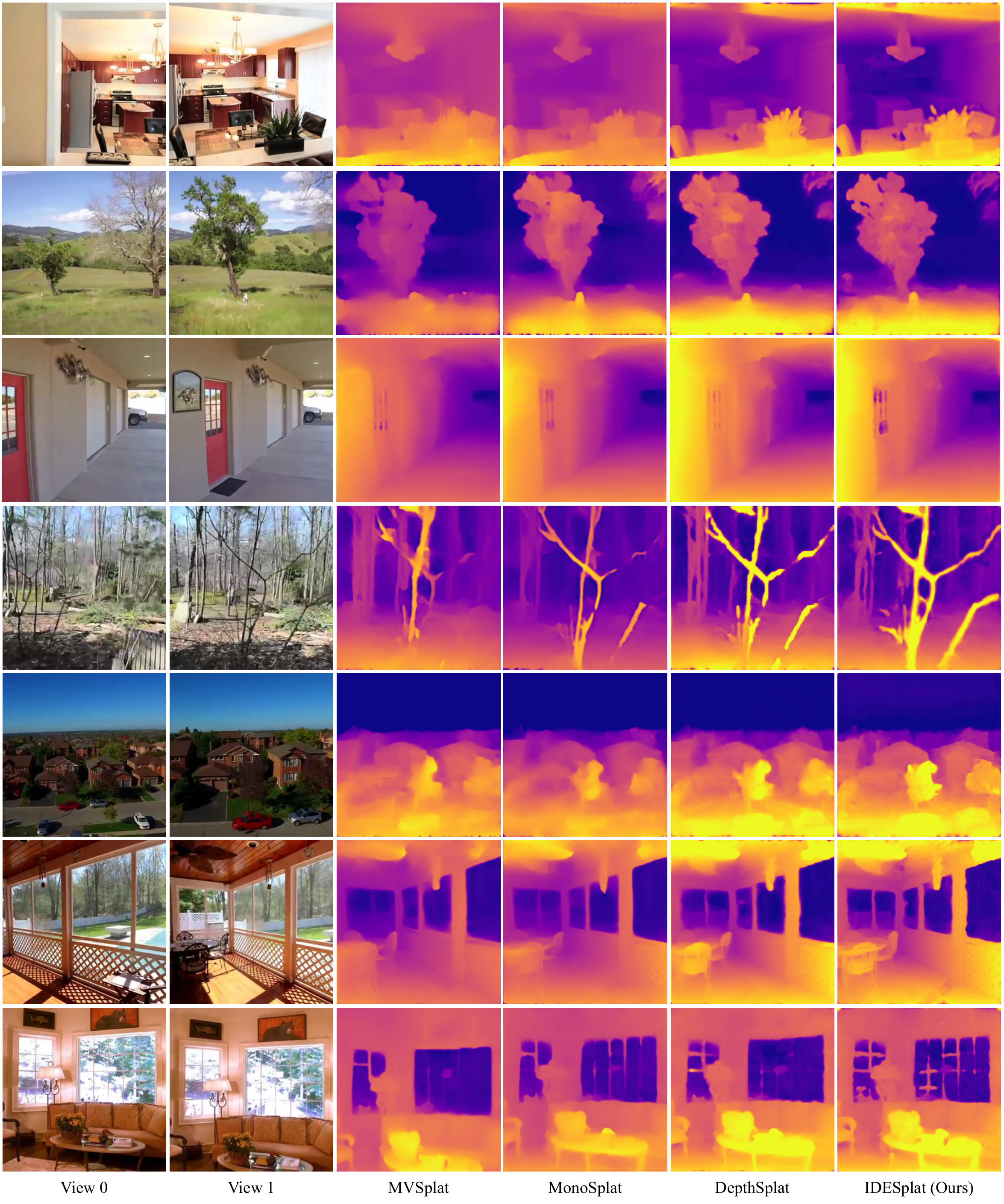}
    \caption{Comparison of depth prediction maps for different models on the RE10K dataset.}
    \label{fig:suppl_re10k_depth_visual}
    \vspace{-3mm}
\end{figure*}

\begin{figure*}
    \centering
    \includegraphics[width=\linewidth]{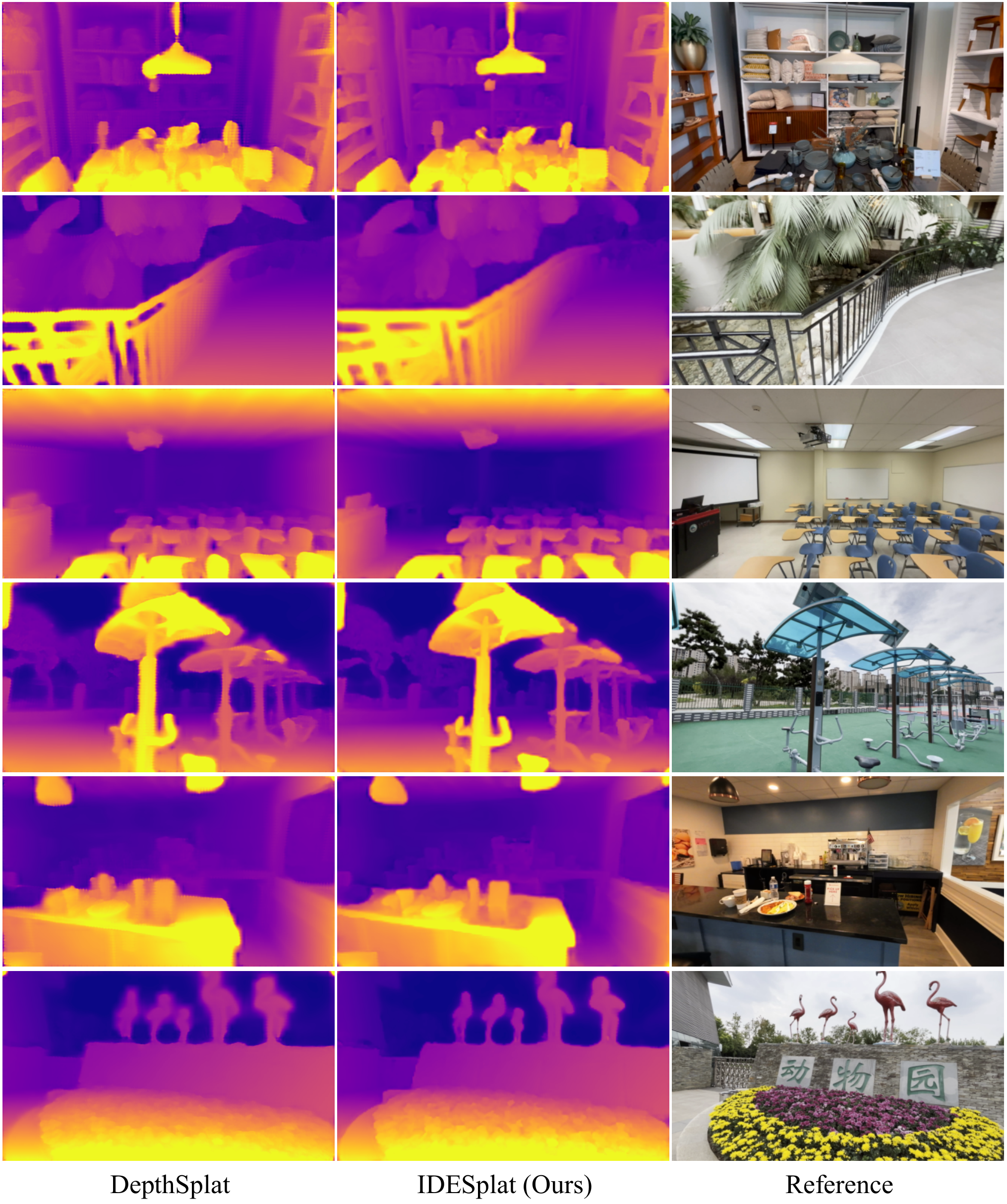}
    \caption{Comparison of depth prediction maps for different models on the DL3DV dataset.}
    \label{fig:suppl_depth_DL3DV}
    \vspace{-3mm}
\end{figure*}

\begin{figure*}
    \centering
    \includegraphics[width=\linewidth]{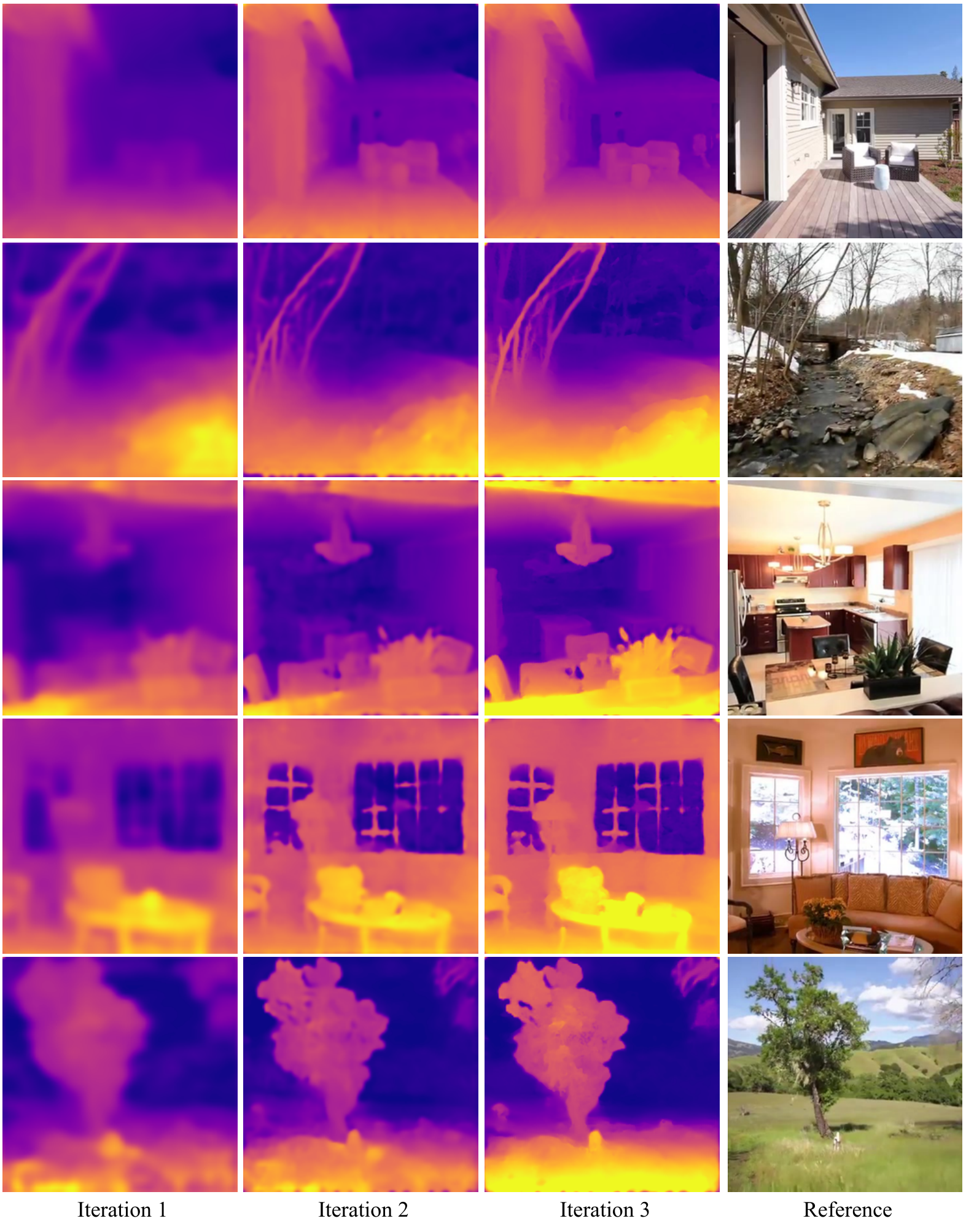}
    \caption{Visualization of intermediate depth prediction maps at different iterations in the IDESplat network.}
    \label{fig:suppl_iteration_depth}
    \vspace{-3mm}
\end{figure*}

We also provide more visual comparison results in this supplementary material, as shown in Fig.~\ref{fig:suppl_re10k_visual} and Fig.~\ref{fig:visual_suppl_DL3DV}. Through these qualitative visual comparisons, it can be observed that our IDESplat outperforms existing methods in novel view synthesis. Even in complex lighting and textured regions, our method achieves better reconstruction results. Furthermore, in Fig.~\ref{fig:suppl_re10k_depth_visual} and Fig.~\ref{fig:suppl_depth_DL3DV}, we provide more qualitative depth map comparisons. The results show that our method significantly improves both the consistency and fine texture details of the depth maps, whether in indoor or outdoor environments, compared to existing methods.
To show how IDESplat refines depth maps over iterations, we compare results visually in Fig.~\ref{fig:suppl_iteration_depth}. Each iteration represents one pass through the Depth Probability Boosting Unit.
We can see that with more steps, the depth map gets better and more detailed. After 3 steps, the model can work at the original image size and the depth map is clearer.
With more iterations, the depth map improves. This leads to more accurate Gaussian centers, which creates a better scene reconstruction.

\end{document}